\documentclass[
twocolumn,
]{ceurart}

\usepackage{listings}
\begin{document}

\copyrightyear{2025}
\copyrightclause{Copyright for this paper by its authors.
  Use permitted under Creative Commons License Attribution 4.0
  International (CC BY 4.0).}

\conference{CLiC-it 2025: Eleventh Italian Conference on Computational Linguistics, September 24 — 26, 2025, Cagliari, Italy}

\title{Doing Things with Words: 
Rethinking Theory of Mind Simulation in Large Language Models}

\author[1]{Agnese Lombardi}[%
email=agnese.lombardi@phd.unipi.it,
url=https://agneselombardi.github.io/,
]
\cormark[1]
\address[1]{CoLing Lab, Department of Philology, Literature and Linguistics, University of Pisa}

\author[1]{Alessandro Lenci}[%
email=alessandro.lenci@unipi.it
]

\cortext[1]{Corresponding author}

\begin{abstract}
    Language is fundamental to human cooperation, facilitating not only the exchange of information but also the coordination of actions through shared interpretations of situational contexts. This study explores whether the Generative Agent-Based Model (GABM) Concordia can effectively model Theory of Mind (ToM) within simulated real-world environments. Specifically, we assess whether this framework successfully simulates ToM abilities and whether GPT-4 can perform tasks by making genuine inferences from social context, rather than relying on linguistic memorization.  
    
    Our findings reveal a critical limitation: GPT-4 frequently fails to select actions based on belief attribution, suggesting that apparent ToM-like abilities observed in previous studies may stem from shallow statistical associations rather than true reasoning. Additionally, the model struggles to generate coherent causal effects from agent actions, exposing difficulties in processing complex social interactions. These results challenge current statements about emergent ToM-like capabilities in LLMs and highlight the need for more rigorous, action-based evaluation frameworks.
\end{abstract}


\maketitle

\section{Introduction}

Language constitutes a fundamental aspect of human cooperative activity, serving not only to describe or assert aspects of a situation but also to actively shape and create situations. Its principal function is communication, which is inherently an interactive process. Through communication, individuals engage in coordinated actions, relying on shared interpretations of their context to align their behaviors and objectives \citep{Winograd2006}.

A notable example is the \textbf{conversation for action}, a structured interaction in which one party (\textbf{Speaker A}) issues a request to another party (\textbf{Speaker B}). This request is understood by both parties as defining specific conditions of satisfaction that outline a prospective course of action for B. Following the initial request, B may respond by accepting (thereby committing to fulfilling the conditions), declining (terminating the conversation), or proposing a counter-offer with modified conditions. Each of these responses opens the possibility for further continuations; for instance, after a counter-offer, A may choose to accept, withdraw the request, or propose an alternative counter-offer in return\citep{winograd86}.

However, all the possibilities available to B are constrained by a specific interpretation of A's utterance. To generate actions that are coherent within a given scenario, agents engaged in communication must accurately interpret natural language, often relying on inferential processes \citep{Grice, Levinson, Sperber, Pinker} or mentalizing abilities to understand others' beliefs, intentions, and access to sentence meaning. This implies that B has a finite set of possible interpretations of A’s utterance, and that each interpretation is associated with a potentially infinite set of possible actions. Thus, different actions may arise depending on how the same utterance is understood.

One reason for this variability is that speakers do not always convey their intended meaning literally. Rather, listeners often need to infer the communicative intent by drawing connections between linguistic meaning and extralinguistic cues, such as the situational context, conventional usage, and past experience.

Let us hypothesize that speaker A’s utterance is a sentence like \emph{Can you open the window?} and that A wants to express an indirect speech act (ISA; \citep{Searle}). Only once B accesses A’s intended meaning can B consider the appropriate set of possible actions to perform. The same principle applies to an utterance like \emph{It is cold here}, which is literally a simple statement about temperature, but can easily be interpreted as an indirect request to turn on the heater. To access the intended meaning, B must use mentalizing abilities to connect the linguistic expression with the situational context (e.g., knowing that a heater is available and that A usually prefers it to be on, etc.). Clearly, the set of actions available to B changes depending on how the utterance is interpreted.

The factor that determines which interpretation is favored by the listener is the accurate inference of the speaker’s beliefs and intentions at the moment the utterance is produced. In other words, they require a \textbf{Theory of Mind} (ToM) and the capacity for ``mentalizing'', that is the ability to reason about others' mental states, to effectively link language with actions within a given situational context. 
A crucial aspect of the ToM involved in communication are \textbf{second-order beliefs}, expressing an agent's mental states about the content of the other agent's mental states (e.g., \textit{John believes that Marks believes that q}). Addressing the formalization and communication of intentions thus necessitates an understanding of language as a form of communicative action. This approach inherently entails the consideration of extralinguistic factors, as demonstrated in studies on multimodal communication \citep{multimodal}, and requires more sophisticated models of situational contexts to comprehensively capture the interplay between language use and interpretation.

Traditionally, the evaluation of Large Language models (LLMs) has largely overlooked the relationship between language and action, instead focusing primarily on the communicative context and dialogue. This omission is, in part, due to the inherent challenges associated with assessing the agentive aspect of language and its connection to actions.

This study proposes the use of the \textbf{Generative Agent-Based Model} (GABM) \textbf{Concordia} \citep{concordia} to embed utterances and narratives within a situational context. The goal is to determine whether reproducing such complex scenarios -- closely resembling real-world environments -- can facilitate the discrimination between intended and literal meanings. Our primary research objective is to assess Theory of Mind (ToM) abilities, operationalized in this experiment as the capacity to infer intended meaning based on extralinguistic factors.

Rather than directly prompting the model to interpret the meaning of an utterance, we ask it to identify the most probable action that the listener would choose, given specific preconditions. This approach is justified by the assumption that each interpretation of an utterance is linked to a set of possible actions.

Our experiment takes into account the overlap between literal and non-literal meanings and the inference processes required for the listener to comprehend the intended meaning of an utterance. In our stimuli, we incorporate utterances that allow for both direct and indirect interpretations. Thus, different actions may arise depending on how the same utterance is understood.

To control for conventional utterance-action associations, we adapt the \textbf{False-Belief task} \citep{Wimmer83} into a novel experimental format. By evaluating action selection rather than meaning comprehension directly, we minimize concerns that the model may have been exposed to the intended meanings during training. Moreover, our task introduces two layers of complexity: first, the model must infer the correct meaning under a false-belief condition; second, it must map that inferred meaning to an appropriate action.

This approach offers several advantages. Following \citet{kim2023fantombenchmarkstresstestingmachine}, we adhere to the two key criteria for a ToM task outlined by \citet{quesque2020}: \textit{non-merging} and \textit{mentalizing}.  

The \textbf{non-merging} criterion requires that evaluation tasks ensure a clear distinction between an agent’s own mental state and that of others. This distinction is often absent in many LLM evaluations, as these models typically process the entire conversation as input, granting them ``omniscient knowledge''. Consequently, it becomes challenging to determine whether a model’s response reflects a character’s belief or results from its comprehensive access to the conversation history. In contrast, our approach explicitly separates the mental states of characters and ensures that their actions are determined solely by their individual knowledge and intentions.  

The \textbf{mentalizing} criterion stipulates that lower-level cognitive processes should not account for successful performance on ToM tasks. If a simpler explanation suffices, it should be preferred over a more complex one when interpreting results. In our framework, we introduce a clear distinction: the speaker's responses and actions can be directly inferred from world-state correlations, whereas the listener's responses and actions necessitate a more intricate mentalizing process. This process requires reasoning about language, context, intentions, beliefs, and desires. To further support this distinction, we present multiple versions of the same narrative, systematically altering agents’ knowledge to encourage diverse interpretations.

Our results reveal a critical limitation: Modeling situational context through real-world simulations is insufficient to elicit ToM-like abilities in the model. Specifically, \textbf{GPT-4 frequently selects actions without appropriately interpreting utterances and the belief context, demonstrating a clear divergence from the ToM capabilities observed in humans}\footnote{Code and dataset available on GitHub: \url{https://github.com/agneselombardi/Concordia_ToM}}.

\section{Related Work}
\subsection{Generative Agent-Based Models}

Generative Agent-Based Models (GABMs) represent a significant departure from traditional agent-based models, which have typically been employed at a relatively high level of abstraction. Moreover, the application of traditional models has been largely confined to specific domains, such as empirical social research \citep{bruch}, market simulations \citep{bonabeau}, and computational sociology \citep{macy}. By contrast, GABMs \citep{gabm-hee, gabm-bubeck, gabm-zhao} enable more precise simulation of behaviors across diverse contexts, leveraging the extensive knowledge embedded in LLMs. These agents not only have a more sophisticated array of cognitive functions for adaptive decision-making but also engage in natural language communication with one another, further enriching their interactive capabilities.

\subsection{Theory of Mind Simulation with Agents}

Theory of Mind (ToM), defined as the ability to infer the beliefs and intentions of others \citep{Villiers}, has been extensively studied in the context of LLMs to assess their capacity for handling complex tasks that require ToM reasoning. A variety of text-based benchmarks, often inspired by established psycholinguistic tests such as the Sally-Anne test \citep{baron-cohen}, have been developed to evaluate this ability. While some findings suggest that LLMs demonstrate remarkable performance on ToM-related tasks \citep{Kosinski_2024}, other studies highlight significant challenges faced by these models in making complex ToM inferences \citep{ullman2023largelanguagemodelsfail}. Consequently, the debate surrounding the extent of LLMs' ToM capabilities remains open. 

Previous works have formalized ToM as agents' knowledge in various contexts, particularly to enhance collaboration in multi-agent reinforcement learning settings \citep{oguntola2023theorymindintrinsicmotivation} and to improve the cooperative behaviors of LLM-based agents through explicit belief modeling \citep{Li_2023}. 
However, these experiments are predominantly conducted in simplified environments, such as the box game task, which differ significantly from the complexities of real-world social scenarios.
On the other hand, previous attempts to model ToM and social interactions have primarily relied on simplified ABMs to simulate developmental settings \citep{tom-abm}.

To the best of our knowledge, our study represents the first attempt to utilize a Generative Agent-Based Model (GABM) to explore: \begin{enumerate}
\item whether LLMs exhibit ToM-like abilities in real-life scenarios and simulations involving pragmatic interpretation, like with ISA.
\item if we can effectively isolate mentalizing from other variables, such as the memorization of linguistic context \citep{wu2024reasoningrecitingexploringcapabilities} and better assess whether a model truly demonstrates ToM capabilities rather than relying on surface-level statistical patterns.
\item whether prompting LLMs with GABM settings leads to more aligned and contextually appropriate outputs.
\item whether adding explicit agents' second-order beliefs and contextual information improves the model's capacity to perform ToM tasks.
    
\end{enumerate}

\section{Concordia}

In Concordia \citep{concordia}, both the model of the environment and the model of individual behaviors are generative. The model responsible for the generation of the environment is called \textbf{Game Master (GM)}.\footnote{The name and the approach reflect the game Dungeons and Dragons, where the Game Master is the player that has the role of storytellling.}

\begin{figure*}
    \centering
    \includegraphics[width=1\linewidth]{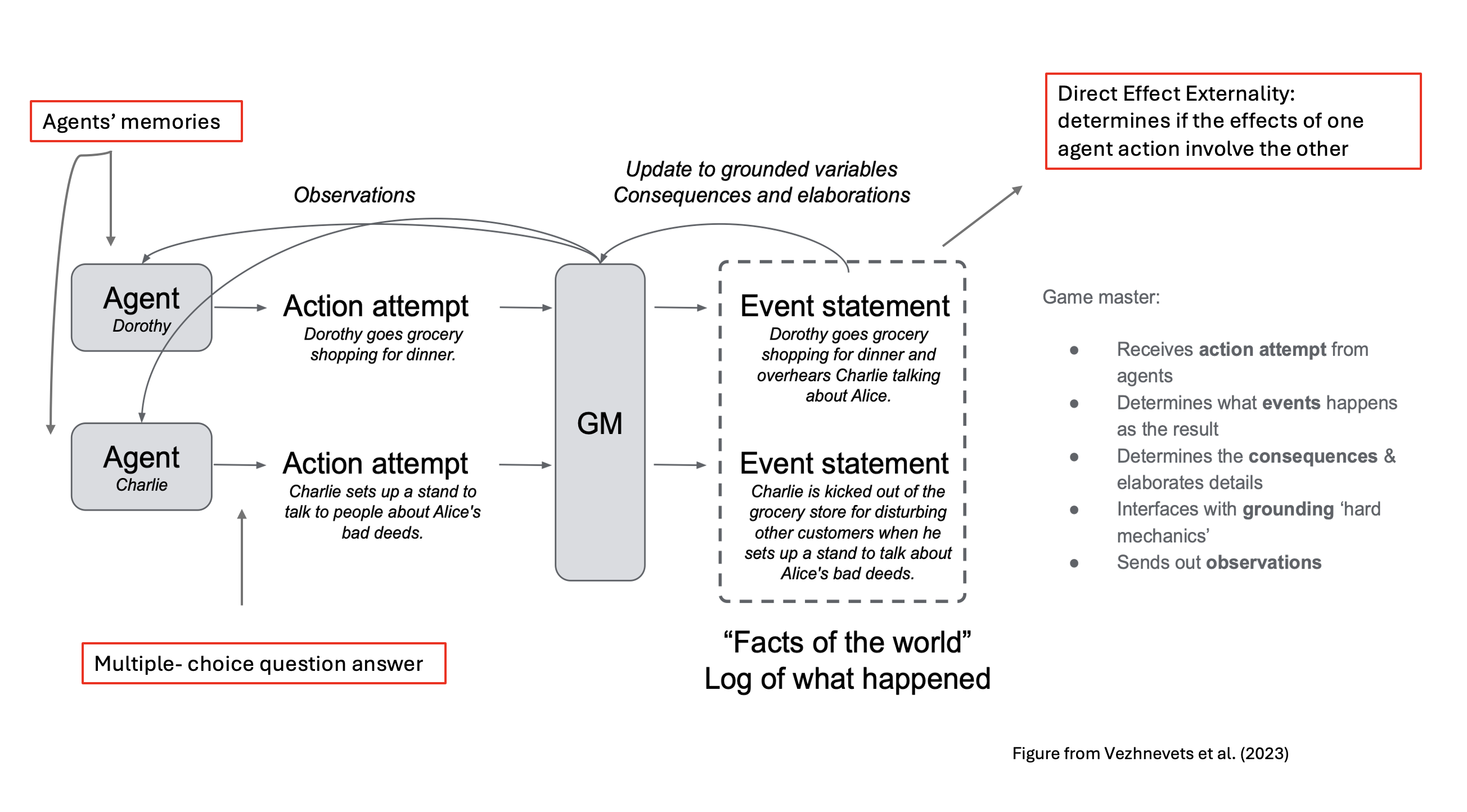}
    \caption{The Game Master mediates between agents and the environment, translating agent actions into environmental observations, while agents adapt their actions based on memory and updated observations.}
    \label{fig: concordia_workflow.png}
\end{figure*}

Figure \ref{fig: concordia_workflow.png} illustrates the structure of the simulation in Concordia. The GM functions as an intermediary between the agents and the environmental dynamics resulting from their actions. Specifically, the GM receives the agents' actions and translates them into corresponding observations, reflecting the environmental effects of those actions. Meanwhile, the agents formulate and execute action strategies informed by their memory and the observations provided by the GM. These observations are subsequently updated to align with changes occurring within the environment.
Observations, actions and event
statements are all English strings.
The GM is also responsible for maintaining and updating grounded variables, advancing the clock and running the episode loop.

In our simulation, agent actions are determined by answering a Multiple-Choice Question Answer (MCQA). The agents' memories encompass all relevant background information necessary for action selection. To enhance coherence, we incorporate a component termed \textit{Direct Effect Externality} following the environment update. This component determines whether the selected actions affect one or more agents and specifies the resulting effects.
This serves as a verification mechanism to ensure that the produced effects on the other player are coherent with the selected action and with the inferred beliefs and desires (that are explicitly codified in the GM memory).

\begin{figure}
    \centering
    \includegraphics[width= 7 cm]{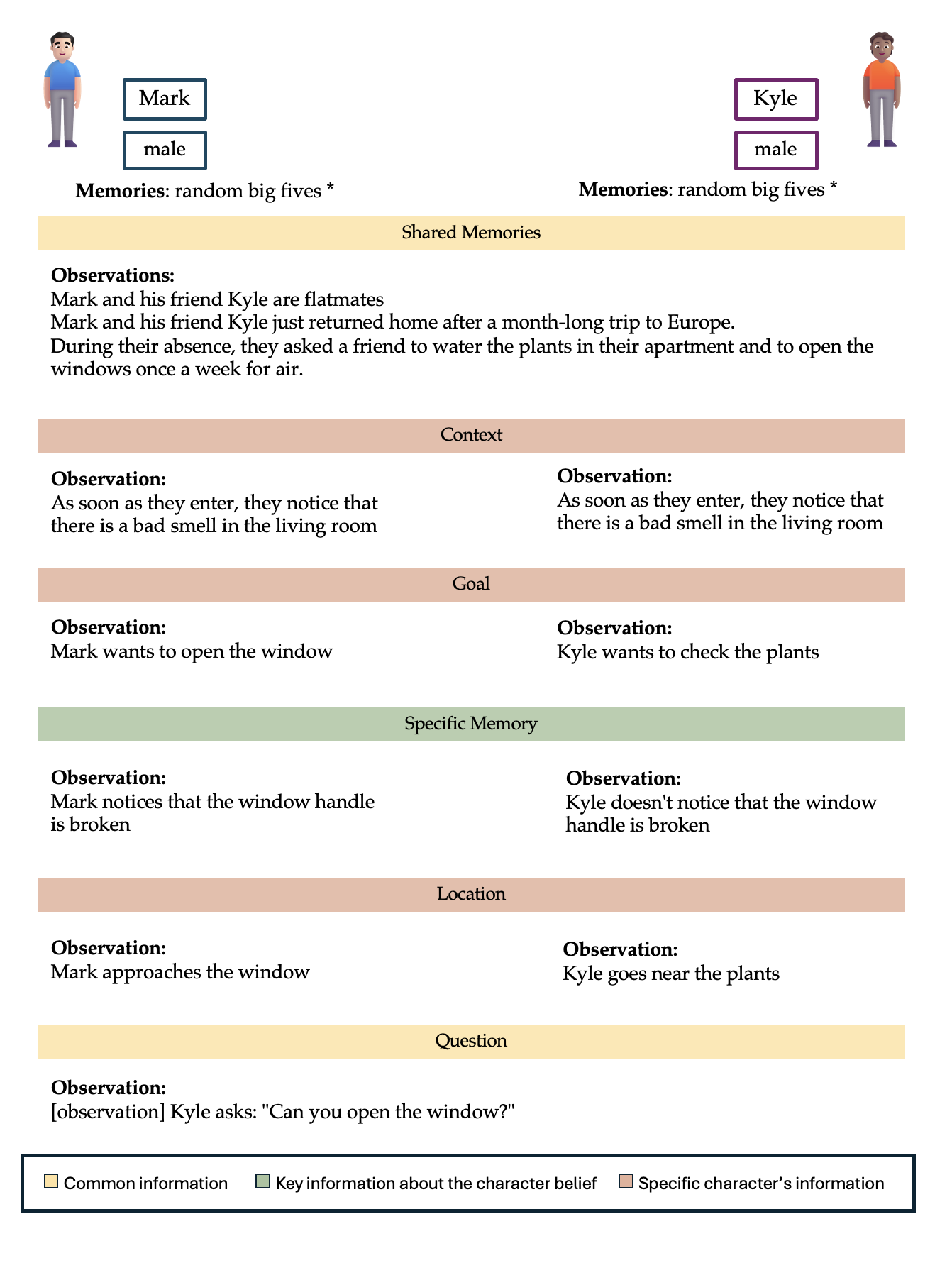}
    \caption{An example of the agents' memory and the type of observation, where the stimulus reproduces an Indirect Request.}
    \label{fig:simulation}
\end{figure}

\section{Simulation}

We generated a total of 200 ToM simulations, grouped into 5 tasks. Each simulation involves two distinct characters, accompanied by a sequence of observations for each of them. The character memory is individually constructed by randomizing the Big Five personality traits \citep{traits}.

The simulation concludes with the final utterance from one of the two characters, which can be interpreted literally or non-literally. This final utterance is constructed to incorporate various pragmatic phenomena that require the use of ToM. Specifically, the utterance can include four types of \textbf{Indirect Speech Acts} (Indirect Requests, Indirect Suggestions, Indirect Declinations, and Indirect Threats) and three forms of \textbf{Verbal Irony} (Sarcasm, Hyperbole, and Rhetorical Questions).\footnote{All stimuli used in this study are manually constructed, with the exception of a subset of indirect requests, which are sourced from \citep{trott-bergen}.}

In each simulation, there is \textbf{shared information} available to both characters as well as \textbf{character-specific memory}, including their goals, locations, and first- and second-order beliefs (see Figure \ref{fig:simulation}). We manipulate the agents' knowledge in a manner analogous to a \textbf{False-Belief task}\footnote{The False-Belief task is a widely used method to investigate ToM \citep{wellman}. It enables a clear distinction between an agent’s true belief and their awareness of another individual’s differing (false) belief.}.
Indeed, the specific memory of the agents is manipulated to evaluate whether the action (response) of the \textbf{listener agent} depends on accurately inferring the beliefs of the other \textbf{speaker agent}, in alignment with ToM. This is achieved by both withholding explicit information about the other agent’s beliefs and providing it to the character.

\begin{figure*}
    \centering
    \includegraphics[width=1\linewidth]{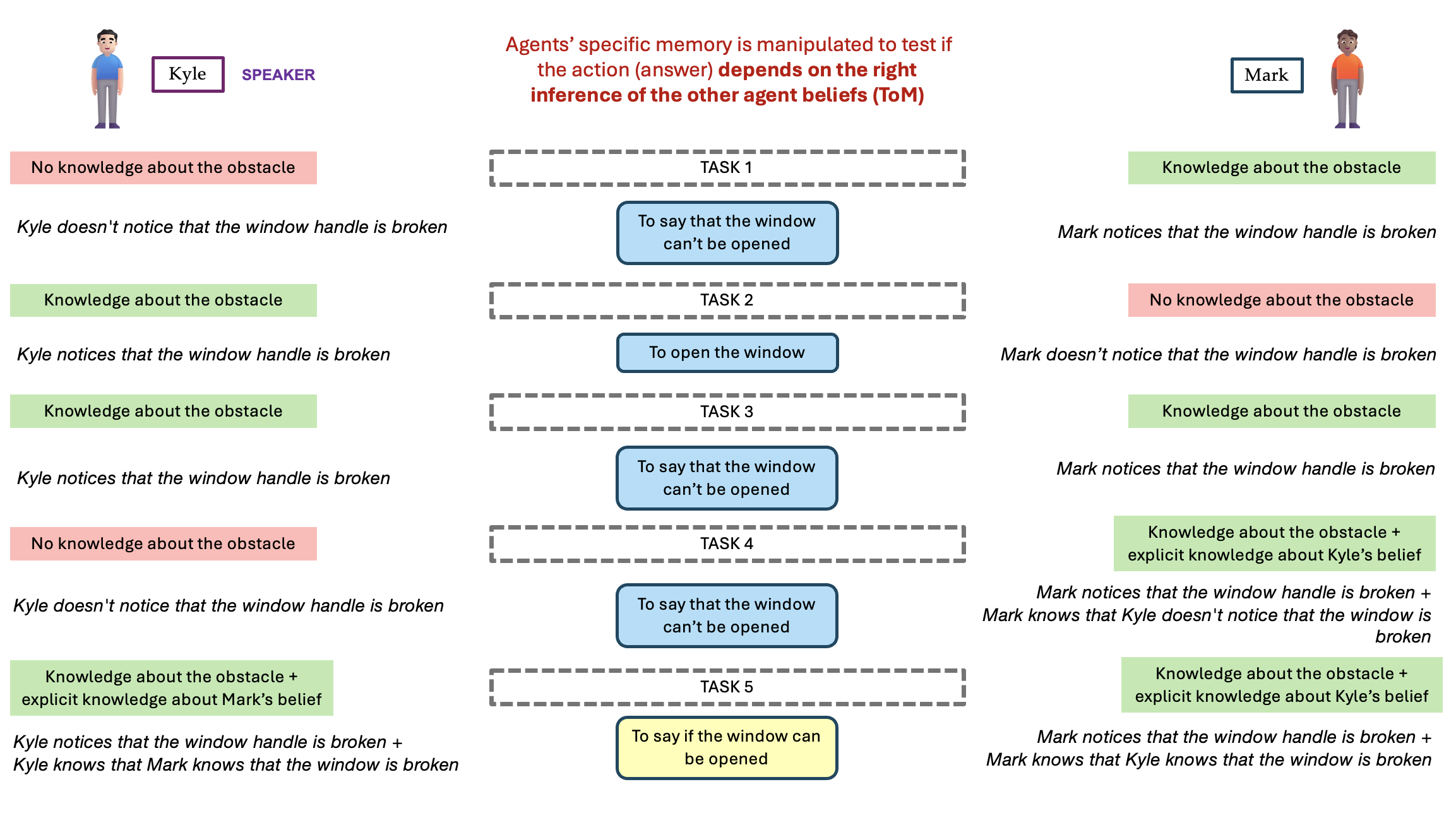}
    \caption{Adjustment of agents' knowledge for the same stimulus across different task designs.Tasks 4 and 5 involve second-order beliefs. The attempted action of the listener agent is determined by the interpretation of the final utterance, which may be understood either literally or non-literally. The action that aligns exclusively with the literal interpretation is highlighted in yellow.}
    \label{fig tasks}
\end{figure*}

The distinct design of each task controls for the agent's beliefs and knowledge regarding the other agent's beliefs. Tasks 1, 2, and 3, take into account only agent's first-order beliefs, whereas Tasks 4 and 5 involve second-order beliefs (Figure \ref{fig tasks}).

In total, there are $8$ stimuli for each linguistic phenomenon, resulting in $40$ stimuli for each task. Ultimately, the objective is to assess whether the selected action by the listener aligns not only with the agent's own intentions and beliefs but also with the resulting consequences in the environment and their impact on the other agent. The generated events are designed to ensure that they account for the beliefs and intentions of both agents.

\subsection{Stimuli}

Since each simulation replicates a false-belief pattern by introducing an obstacle to the indirect interpretation of the final utterance and manipulating agents' awareness of this obstacle, we designed $5$ versions of the simulation (Figure \ref{fig tasks}). In these tasks, i.) the agents' knowledge of the obstacle is systematically varied through the information stored in their specific memory, and ii.) knowledge variation determines whether the speaker's sentence is interpreted literally or not, iii.) which in turn prompts a certain action by the listener. This allows us to control whether \textbf{the action produced by the listener is consistent with the most likely interpretation of the speaker's sentence, given the agents' knowledge in the scenario}.
Thus, both interpretation and action are contingent upon the ability to infer the beliefs and desires of the other agent. As illustrated in Figure \ref{fig tasks}, given a test item represented by the sentence $S$: \textit{Can you open the window?}, we have the following tasks:

\begin{itemize}
\item[]\textbf{Task 1} -- \textbf{the speaker is unaware of the obstacle (\textit{The handle is broken}), while the listener is aware of it}. The listener is expected to interpret S with the non-literal meaning (i.e., indirect request), and thus the most likely action would be to inform the speaker that the window cannot be opened.\footnote{The intended meaning here is \textit{I am asking you to open the window}, reflecting the speaker's desire (\textit{D}: to open the window) and belief (\textit{B}: the window is not broken). However, the listener knows that the window handle is broken and, therefore, that the window cannot be opened. Therefore, the listener holds a different belief \textit{B1} from that of the speaker.  
If the listener lacks knowledge about the speaker's beliefs and desires, the interpretation of \textit{S} may default to a non-literal meaning. This phenomenon aligns with findings from psycholinguistic experiments, where default interpretations often prevail when they are more conventionalized than the literal ones \citep{Gibbs, Gibbs_83}.}

\item[] \textbf{Task 2} -- \textbf{The agents' beliefs are reversed compared to those in Task 1}.\footnote{The listener lacks knowledge of the obstacle and thus holds belief \textit{B}, while the speaker holds belief \textit{B1} (cf. previous footnote).} In this scenario, the default interpretation is the non-literal one, but the listener is expected to attempt to open the window based on its own belief that the window can be opened.

\item[]\textbf{Task 3} -- \textbf{Both agents are aware of the obstacle (\emph{The handle is broken}), but there is no explicit knowledge of the other agent's belief}. The expected listener's action is to inform the speaker that the window cannot be opened, like in Task 1. This scenario becomes particularly informative when compared to Task 5 below, where both agents are explicitly provided with second-order beliefs. Even in agent-based models where agents acquire information about the situation and context, it is essential to possess knowledge of the other agent's beliefs in order to select actions that are coherent with the situational context.

\item \textbf{Task 4} and \textbf{Task 5} -- \textbf{They are extended versions of Task 1 and Task 3, respectively, incorporating \emph{second-order beliefs}}. In Task 5, the interpretation of \textit{S} is expected to be literal: Since both agents are aware that the handle is broken, the intended meaning of \textit{S} should be \textit{I want to know if the window can be opened despite the broken handle.}
\end{itemize}

This manipulation of character knowledge allows us to investigate whether and how the interpretation of an utterance varies depending on the belief states of the speaker and the listener. In the first three tasks, the model is provided only with character-specific knowledge, simulating real-world conversational dynamics in which speakers must infer others’ mental states based on context. Here, the listener interprets the utterance based solely on their own knowledge, and any correct or incorrect understanding of the intended meaning arises from inferences about the speaker's beliefs.
In contrast, Tasks 4 and 5 introduce \textbf{explicit representations of others’ beliefs in the form of second-order beliefs} (e.g., \textit{Mark knows that Kyle knows that the window is broken}). In these tasks, the listener has access not only to their own knowledge but also to the knowledge state of the speaker. Consequently, action selection depends on i.) the model's capacity to reason over second-order beliefs and ii.) its integration of this information with its own knowledge. This setup allows us to distinguish between first-order and second-order ToM capabilities in model behavior. 

\section{Experiments}
The first phase of our experiment is formulated as a Multi-Choice Question Answering (MCQA) problem, in which the model is provided with an agent's memories and observations, followed by a question regarding the agent's likely next action, along with four possible answer choices (see Figure \ref{fig:action}, Appendix \ref{sec:app-sim}). Concordia performs a separate API call for each agent, ensuring that it generates an independent response. The four answer choices correspond to the possible responses derived from different simulation scenarios (see Figure \ref{fig:simulation}).
At the time of the experiments, Concordia had not been adapted for open-source models yet. Therefore, we opted for GPT-4o-mini,\footnote{Prompted 22 November 2024} which has demonstrated state-of-the-art performances across a wide range of ToM tasks.

In the second phase, the GM processes all actions performed by the agents, along with a summary of each agent’s situational context. This information is used to prompt the model using a Chain of Thought approach \citep{wei2023chainofthoughtpromptingelicitsreasoning}. First, the model generates an event statement that updates the environment to reflect the consequences of the performed action -- effectively logging what has occurred (Figure \ref{fig:event}, Appendix \ref{appendix}). Then, the model evaluates whether the action has an impact on the agents and determines the nature of this impact as part of the Direct Effect Externality component. If the event directly affects an agent, both known and unknown effects are generated. Agents' intentions and actions are integrated by the GM within the prompting phase that queries for effects, requiring the model to consider multiple perspectives simultaneously to generate the appropriate outcomes (Figure \ref{fig:dee}, Appendix \ref{appendix}).\footnote{
All memories, prompts, and relevant information are systematically stored in HTML files for documentation and analysis. HTML versions are accessible through the GitHub link.}

\subsection{Evaluation}


To evaluate whether the attempted actions of each agent align with their intentions in the MCQA task, we extracted the generated text for each agent from the HTML files and we compared it with the expected response for that task.

For the evaluation of generated text from the Direct Effect Externality component, we extracted relevant information and additionally prompt GPT-4o-mini to assess the coherence of the effect with the agent’s action and scenario. This process yields the following evaluation template (see Appendix \ref{appendix}) for each agent:  \textbf{Scenario (summary of agent's observation and belief}) + \textbf{attempted action of agent X} + \textbf{Known and/or Unknown effect} + \textbf{coherence rating} (on a scale from 1 to 5, generated by the model).\footnote{When the model determines that there are no direct effects on the agents, it must assign a coherence rating of 0. This ensures that the evaluation framework accurately distinguishes between scenarios where actions produce meaningful consequences and those where no direct impact occurs.}  
This structured approach ensures a systematic assessment of how well the predicted effects align with the agent’s intended actions within the given scenario.

The use of ``LLM-as-a-Judge'', where
LLMs are employed as evaluators for complex tasks, has been shown to be a reliable assessment method \cite{Gu:etal:2025}. Thus, we employ this method to assess the model’s ability to connect actions to social context and to cross-check the coherence it attributes to the effects it generates. Specifically, in the Direct Effect Externality component, a \textbf{Chain-of-Thought} (CoT) is generated based on the event statement produced by the GM after the attempted action -- this statement serves as a summary of the effects that the action produces. However, in our evaluation template, we compare coherence against the initial scenario summary that we originally provided to the model.  
This way, we determine whether, at the end of the cycle, the effect on the agent remains truly coherent with the given scenario and the agent's beliefs, rather than merely aligning with additional effects generated by the model itself.

Following this automated evaluation, two different expert annotators checked the assigned ratings to verify their accuracy and to ensure that the consequences are meaningfully related to the corresponding actions and scenarios. Meanwhile, the assessment of ToM capabilities is derived from the MCQA task.

\section{Results and Discussion}

\begin{figure*}
    \centering
    \includegraphics[width=1\linewidth]{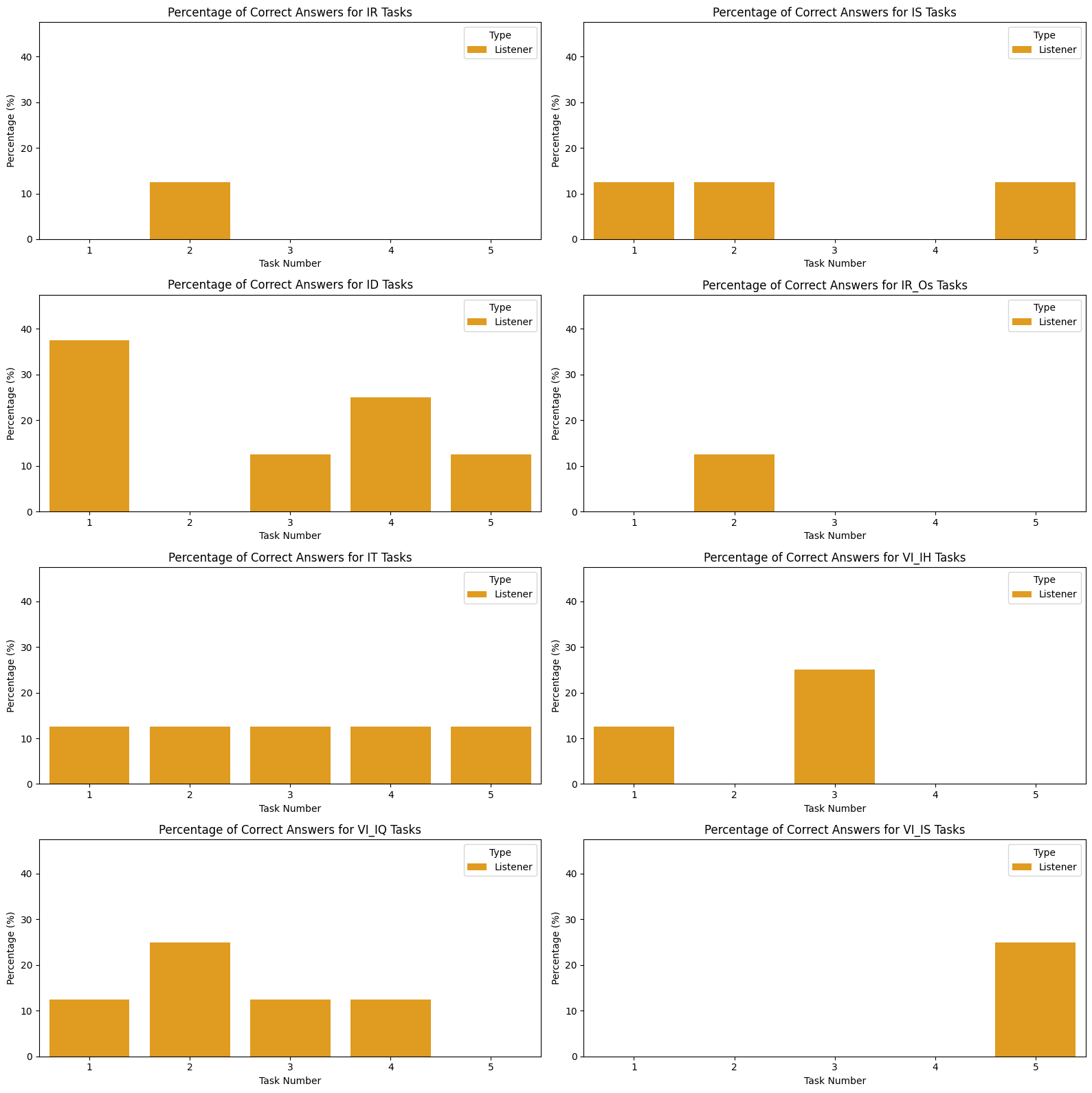}
    \caption{Percentage of correct answers for each task. Orange bars represent the listener, whose correct response varies depending on the scenario.\\
    IR: Indirect Requests; IS: Indirect Suggestions; ID: Indirect Declinations; IR-Os: Indirect Requests extracted from \citet{trott-bergen}; IT: Indirect Threats; VI-IH: Verbal Irony, Indirect Hyperbole; VI-IQ: Verbal Irony, Rhetorical Questions; VI-IS: Verbal Irony, Sarcasm}
    \label{fig:output}
\end{figure*}

\subsection{Actions and Theory of Mind}
\label{actions}

In the initial phase of our experiment, we aim to utilize GPT-4o-mini to replicate ToM-like abilities while simultaneously assessing its capacity to perform ToM tasks within a simulated real-life scenario. Our objective is to determine whether this approach enables an independent evaluation of ToM capabilities, separate from the influence of linguistic context. Then, we seek to determine whether incorporating explicit representations of agents' beliefs enhances the model's performance on ToM tasks. Additionally, we aim to explore potential differences in the model's handling of first-order versus second-order ToM beliefs.

Figure \ref{fig:output} illustrates the percentage of correctly selected actions for each task and linguistic phenomenon. The consistently low accuracy observed across tasks and linguistic phenomena indicates that the model struggles to select context-appropriate actions, and by extension, to derive the correct interpretation of utterances through ToM-like reasoning. This finding is particularly noteworthy when considered within the broader context of recent ToM-related studies, many of which—especially those focusing on OpenAI models—have suggested a more optimistic picture of such capabilities \citep{Kosinski_2024}.

No clear pattern emerges across tasks, nor is there a significant difference between first-order and second-order belief tasks. This lack of systematic variation suggests that the model does not exhibit ToM-like abilities, as its responses do not consistently reflect any process similar to belief attribution or true mental inferencing. Therefore, \textbf{the GABM is not able to use either first- or second-order beliefs -- despite the fact that these have been explicitly given to it -- to interpret the speaker's sentence consistently with the knowledge setting in the scenarios}. 



\begin{figure*}
    \centering
    \includegraphics[width=1\linewidth]{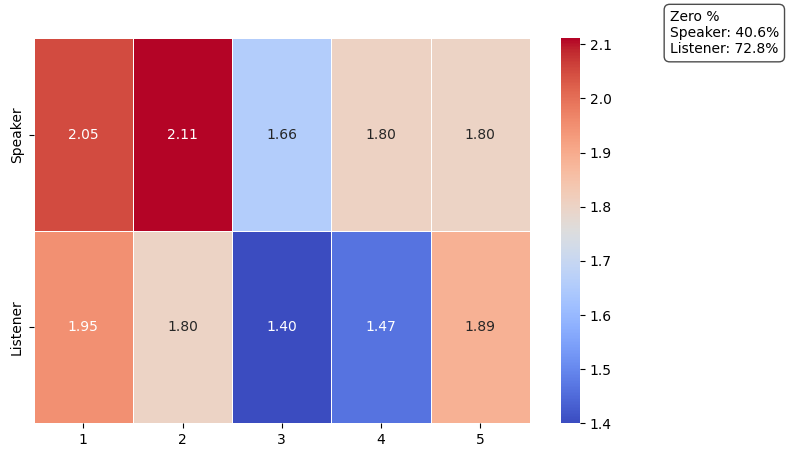}
    \caption{Ratings are assigned by prompting the model to evaluate the coherence of the generated effects on agents in relation to the given context and attempted actions. When no effect on the agents is present, the model must assign a rating of 0 -- the probability of such cases is displayed in the box at the top right. For all other instances where an effect is generated, the assigned coherence ratings for both the speaker and the listener must fall within the range of 1 to 5.}
    \label{fig:ratings}
\end{figure*}

\subsection{Causal-Effect Coherence}
\label{random}

In this analysis, we aim to investigate whether the GABM setting leads to more contextually aligned and appropriate outputs. We compared the effects generated by the model in response to agent actions with both the predefined scenario and the beliefs assigned to the agents. We then assessed whether the model itself considers these effects coherent by assigning a coherence rating on a scale from $1$ to $5$.  Following this automated evaluation, we manually reviewed the model’s ratings to assess their accuracy. 



As illustrated in Figure \ref{fig:ratings}, the model assigns notably low coherence ratings to effects that it itself generates, with a maximum average rating of 2.11 on a scale from 1 to 5.
The observed discrepancy between the selected action and the generated consequences highlights the model’s difficulty in integrating situational context with utterance interpretation in a coherent manner. The CoT reasoning often reflects limited contextual awareness, focusing primarily on short-range dependencies rather than engaging in the broader reasoning processes necessary to produce coherent cause-effect relationships.
To better illustrate this contrast, we included the model’s self-evaluation of its outputs and compared these judgments with those of human annotators. This comparison underscores a critical distinction: during generation (i.e., in the CoT), the model is required to actively infer and reason about the situational context in order to produce a logically coherent narrative. However, when evaluating its own output, the model can rely on the full textual context and potentially draw on patterns and examples present in its training data. Interestingly, in this evaluative mode, the model's coherence judgments align more closely with human assessments—likely because the task resembles familiar forms of pattern recognition, rather than the more demanding process of causal reasoning required during generation.

\section{Conclusion}

Our objective was to utilize the Generative Agent-Based Model Concordia to reframe ToM tasks and investigate whether mentalizing abilities could be isolated from other confounding variables typically present in prompting-based evaluations. Specifically, we aimed to reproduce a standard False-Belief task within a complex social simulation. To achieve this, we carefully designed stimuli involving uncommon social situations to determine whether modeling a rich situational context and assigning explicitly to the model first- and second-order beliefs would aid it in making the correct inferences and producing an action consistent with the knowledge scenario.  

The results presented in Section \ref{actions} underscore a growing concern in the Theory of Mind (ToM) research community: \textbf{The challenge of designing tasks that effectively isolate ToM-like abilities in LLM from confounding variables}. Our findings raise important questions about the mechanisms driving ToM-like performance in state-of-the-art LLMs and the true nature of their so-called emergent abilities.
For example, while the False-Belief task remains a widely used and valuable benchmark for testing ToM, it is also a well-known paradigm likely to appear in post-training data. This raises legitimate concerns about whether models are genuinely reasoning about beliefs or simply learning how to solve familiar tasks through exposure. Furthermore, although the False-Belief task is well-established in human cognitive testing, the conditions under which it is administered differ significantly from those we can replicate in computational models. 
While we maintain that it remains a useful tool for evaluating ToM-like capabilities, we argue that \textbf{it should be supplemented with additional constraints and more indirect testing methods -- such as connecting utterance interpretation with action selection}, as we do in our work -- rather than relying solely on metalinguistic judgments. Our results lend support to the memorization hypothesis, suggesting that current LLMs may not truly reason about propositional attitudes but instead exploit learned statistical patterns present in their training data.

Additionally, the model does not consistently select coherent effects in response to actions, indicating that we are still far from developing frameworks that accurately model complex social scenarios. However, employing these agent-based simulations as evaluation methods represents a promising research direction.  
It is reasonable to conclude that LLMs remain far from producing fully aligned and contextually coherent outputs in tasks requiring deep social reasoning.
We conclude that to isolate ``mentalizing'' processes, we should rely on more complex scenarios, focusing on assessing \textit{functional ToM} rather than merely \textit{literal ToM} \citep{riemer2025positiontheorymindbenchmarks}. 

\section{Limitations}

 This study has several limitations. First, it relies heavily on the model’s self-evaluation, introducing a risk of circular reasoning. 
 
 Human evaluation was limited to two annotators, restricting claims about inter-annotator reliability. Additionally, we used pre-existing components of Concordia rather than developing tools specifically designed for ToM assessment. Our analysis focused solely on GPT-4o-mini, limiting generalizability across models. Finally, we evaluated outputs only, without investigating the internal mechanisms underlying the model's ToM-related reasoning.


\bibliography{custom-2}


\appendix
\section{Appendix}
\vspace{5pt}

\subsection{Simulation display}
\label{sec:app-sim}

At the conclusion of the simulation, all relevant information is collected within the Game Master (GM), allowing us to retrieve segments of the Chain of Thought (CoT) used by the model to determine both the event statement and its effects on the agents. While this framework offers a range of possibilities for modeling social situations, we specifically chose to replicate simple false-belief tasks using Concordia to evaluate whether mentalizing processes could be effectively isolated and to assess whether enriching the social context enhances the emergence of ToM-like abilities.  

\begin{figure}
    \centering
    \includegraphics[width=1\linewidth]{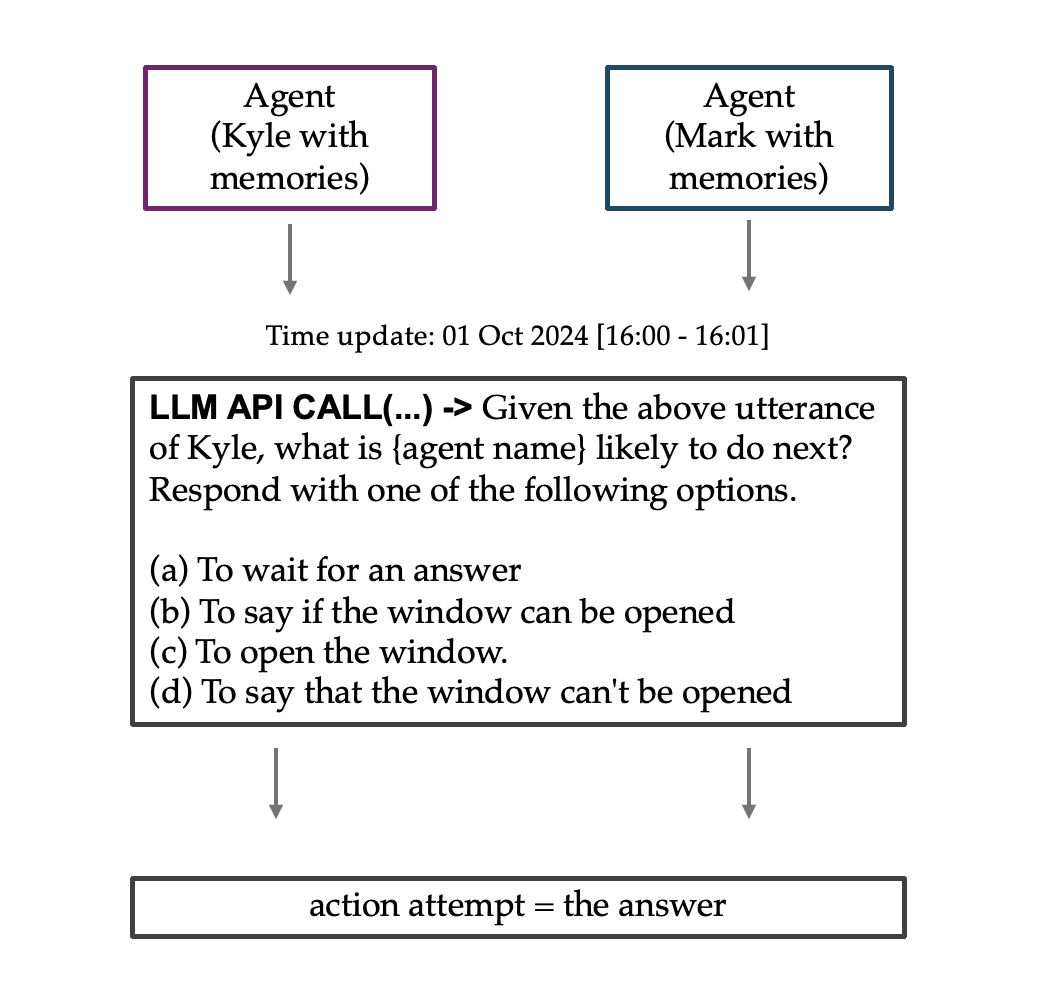}
    \caption{API call to LLM reproducing a Multi-Choice Question Answering task}
    \label{fig:action}
\end{figure}

To achieve this, we implemented two distinct evaluation tasks. First, we employed a Multiple-Choice Question Answering (MCQA) task, in which the model had to select an agent's actions based on their desires and beliefs (Figure \ref{fig:action}. Subsequently, we shifted our focus to assessing the general coherence of the model’s generated actions within the social context. This involved evaluating the model’s ability to utilize CoT reasoning to produce meaningful event statements and generate coherent effects of events on agents.  

\begin{figure*}
    \centering
    \includegraphics[width=1\linewidth]{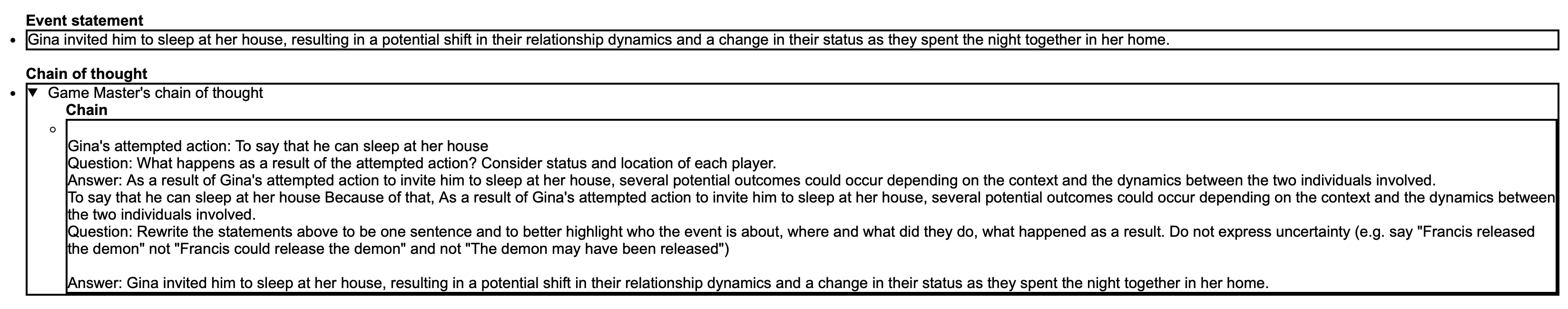}
    \caption{After the GM has received the agent attempted action, it generates a Chain of Thought to determine which events the action caused.}
    \label{fig:event}
\end{figure*}

\begin{figure*}
    \centering
    \includegraphics[width=1\linewidth]{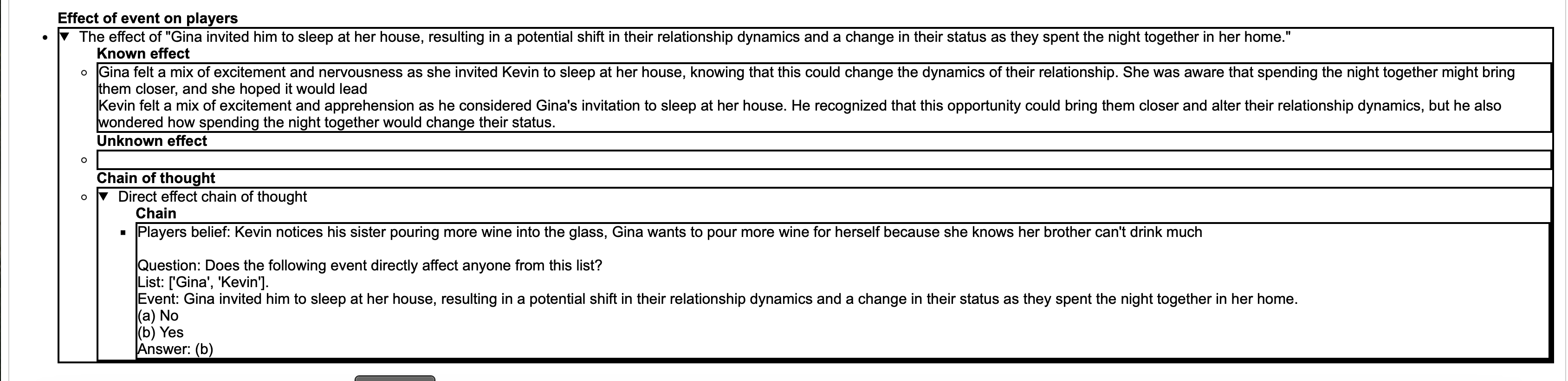}
    \caption{How the Direct Effect Externality component results appear after they have been stored in HTML format}
    \label{fig:dee}
\end{figure*}

In the GABM setting, the model must retrieve previous information to determine the correct effect, yet in some cases, it appears to rely only on the most recent portion of text. This issue is evident in Figure \ref{fig:dee}, where, despite the CoT explicitly containing the player’s belief that the agent notices his sister, this information is lost during the prior CoT steps that summarize observations and actions into an event statement (Figure \ref{fig:event}). The event statement represents a generalized effect of an agent’s action in the environment and is sent back to agents as an observation. It serves as the basis for evaluating whether an action has an effect on the agents themselves.  

Due to this loss of information, the generated effect can sometimes become entirely incoherent with the initial context. This misalignment is reflected in the model’s own coherence ratings, which capture the inconsistency between the intended effect and the final output.

Figure \ref{fig:event} presents an example of an event statement generated based on the attempted action of one of the agents. Figure \ref{fig:dee} illustrates the subsequent process of determining the effects of the action on the agent, considering both the action itself and the event statement. For clarity, we chose to highlight two of the most controversial examples in this discussion.

\subsection{Evaluation Details}
\label{appendix}

To evaluate the coherence of model-generated text in relation to the scenario and the agents' attempted actions, we employed the following assessment template. This template was also used to verify the model’s ratings and determine their alignment with our own judgments.
Template is based on that created by \citet{wu-etal-2024-rethinking}:
\begin{quote}
    We request your evaluation of the AI model’s response in relation to the given scenario. Specifically, consider the scenario involving two agents and their beliefs, assessing whether the model-generated effects align coherently with the agents’ actions and context.

    Evaluate the response based on the following criteria:

    Social Understanding – Does the model grasp the social dynamics and pragmatic nuances of the scenario?\\
    Appropriateness – Is the response contextually relevant and suitable for the scenario?\\
    Insightfulness – Does the answer demonstrate a deep understanding of intentions, implicature, deceit, irony, sarcasm, humor, metaphor, etc.?\\
    Completeness – How well does the response capture the essential elements of the scenario?\\
    Agentivity – Is the model's response coherent with the agents’ attempted actions?
    
    Scoring:
    Assign a score from 1 to 5 for each category.
    Compute a final rating based on these scores.
    If no effect is provided, assign 0.
    Output only a single numeric value representing the final rating (1–5).
\end{quote}



\end{document}